# HGNet: High-Order Spatial Awareness Hypergraph and Multi-Scale Context Attention Network for Colorectal Polyp Detection

Xiaofang Liu[a], Lingling Sun[b], Xuqing Zhang[b], Yuannong Ye[b], Bin zhao[b,*]

[a]*School of Medical Imaging, Guizhou Medical University, Guiyang 550004, China.*

[b]*School of Biology and Engineering, Guizhou Medical University, Guiyang 550004, China.*

**Abstract**

Colorectal cancer (CRC) is closely linked to the malignant transformation of colorectal polyps, making early detection essential. However, current models struggle with detecting small lesions, accurately localizing boundaries, and providing interpretable decisions. To address these issues, we propose HGNet, which integrates High-Order Spatial Awareness Hypergraph and Multi-Scale Context Attention. Key innovations include: (1) an Efficient Multi-Scale Context Attention (EMCA) module to enhance lesion feature representation and boundary modeling; (2) the deployment of a spatial hypergraph convolution module before the detection head to capture higher-order spatial relationships between nodes; (3) the application of transfer learning to address the scarcity of medical image data; and (4) Eigen Class Activation Map (Eigen-CAM) for decision visualization. Experimental results show that HGNet achieves 94% accuracy, 90.6% recall, and 90% mAP@0.5, significantly improving small lesion differentiation and clinical interpretability. The source code will be made publicly available upon publication of this paper.

## 1. Introduction

According to the 2022 Global Cancer Statistics from the International Agency for Research on Cancer (IARC), CRC accounts for 9.6% of new cancer cases worldwide, ranking third in incidence. It also accounts for 9.3% of cancer-related deaths, ranking second globally [1]. Every year, around two million new CRC cases are reported, leading to an estimated 800,000 deaths [2]. Both CRC incidence and mortality rates remain high and continue to rise [3]. The 2024 American Cancer Society (ACS) report further reveals that in the United States, CRC ranks fourth in incidence at 7.64%, and second in mortality at 8.67% [4]. These statistics highlight the ongoing global burden of CRC and the critical need for effective screening strategies. In addition, research shows that nearly 90% of CRC cases develop from colorectal polyps [5]. Early removal of polyps can reduce CRC mortality by 70% [6], emphasizing the necessity of polyp screening.

Colonoscopy remains the gold standard for detecting and removing polyps [7,8]. Although advancements in endoscopic technology have enhanced detection accuracy [9], colonoscopy has been shown to reduce CRC incidence by 52% and mortality by 62% [10]. However, factors such as suboptimal lighting, the varied morphology of polyps, and differences in physician expertise contribute to a detection rate of only 78%, resulting in a significant number of polyps being missed [2,11].

In recent years, deep learning technologies accelerated by GPUs have shown significant promise in improving polyp detection. AI algorithms have demonstrated enhanced detection of small polyps, with their effectiveness validated in trials [12,13].

[*] Corresponding author.
E-mail:.address:,scott84@sina.com (B.Zhao)

However, several challenges remain. First, traditional methods rely heavily on manually designed features, making the development of end-to-end frameworks difficult. Second, CNN-based single-stage and two-stage models struggle to capture multi-scale features efficiently, often missing small polyps. Third, the lack of interpretability in AI model predictions limits clinical application.

To overcome these challenges, we introduce the High-Order Spatial Awareness Hypergraph and Multi-Scale Context Attention Network (HGNet). This model is built upon an enhanced YOLOv11 architecture and integrates the Eigen Class Activation Map (Eigen-CAM) to improve interpretability. Our experiments on three benchmark datasets show that HGNet outperforms most existing models.

The main contributions of this work are as follows:

(1) We proposed the Efficient Multi-Scale Context Attention (EMCA) module, which combines Efficient Multi-scale Attention (EMA) [14] and Context Anchor Attention (CAA) [15] to refine the extraction of both multi-scale contextual and local features for small polyps.

(2) We introduced the spatial hypergraph convolution (HyperConv) method, which captures higher-order spatial relationships in lesion regions, enhancing the representation of complex polyp geometries.

(3) We applied a transfer learning strategy to mitigate the impact of limited public data on model performance.

The structure of this paper is as follows: Section 2 reviews related research on colorectal polyp detection; Section 3 outlines the HGNet architecture and its loss functions; Section 4 introduces the experimental datasets, configurations, and evaluation metrics; Section 5 discusses the experimental results and evaluates the model's strengths and weaknesses; Section 6 concludes the study and proposes future research directions.

## 2. Related Work

In recent years, with the advancement of AI and Machine Learning technologies, deep learning has made remarkable progress in assisting clinicians with diagnoses, utilizing its powerful learning capacity to improve the accuracy of polyp detection [16]. Current polyp detection algorithms primarily include three types: traditional object detection algorithms, two-stage object detection algorithms and single-stage object detection algorithms.

### *2.1. Traditional object detection method*

Traditional polyp detection methods generate candidate regions using sliding windows, relying on manually designed shape, color, and texture features, and classifiers such as SVM or random forests.

Early studies focused on shape features. Summers et al. applied local curvature analysis to identify polyp regions in CT colonography images [17]. Later, methods such as curvature and sphericity were introduced, but kernel-based estimation often led to biases. Zhu et al. improved accuracy by using the Knutsson mapping method [18]. For color features, Karkanis et al. applied color wavelet transformations [19]. Billah et al. combined these with shallow CNN layers [20], while Alexandre et al. focused on pixel-based extraction [21]. For texture, Ameling et al. combined Gray-Level Co-occurrence Matrices (GLCM) with Local Binary Patterns (LBP) [22]. Ren et al. proposed combining shape indices (SI) with multi-scale filters for radiomic feature extraction [23].

Traditional polyp detection algorithms are limited by high computational complexity and poor generalization in clinical scenarios and varying morphologies.

### *2.2. Two-stage object detection method*

Two-stage object detection methods first generate candidate regions before performing classification and localization.

To overcome limitations of manual feature extraction, Byrne et al. developed an end-to-end deep convolutional neural network (DCNN) based on the Inception architecture [24]. However, Su et al proposed deep neural networks (DNNs) often exhibit reduced robustness and higher false-positive rates [25]. Qadir et al. addressed this issue by integrating temporal dependencies into a CNN-based detector, thereby enhancing detection accuracy [26]. To improve region proposal efficiency, Ren et al. introduced the Region Proposal Network (RPN) within Fast R-CNN, significantly accelerating detection speed through shared convolutional features [27]. Based upon this work, Chen et al. incorporated contrast enhancement and attention mechanisms into Faster R-CNN to refine detection accuracy [28]. Qian et al. mitigated specular reflections in colonoscopy images by converting the images into the Hue-Saturation-Value (HSV) color space [29]. Kun et al. integrated ResNeSt and a Side-Aware Boundary Localization (SABL) module into Faster R-CNN, reducing false positives [30]. Jiang et al. improved accuracy with Enhanced CenterNet and Box-assisted Contrastive Learning (BCL) [31].

Despite achieving high accuracy, two-stage detection models require extended training periods, posing challenges for real-time clinical applications.

### *2.3. Single-stage object detection method*

Single-stage detection models are effective in classifying and localizing targets, leading to improved detection rates. Liu et al. introduced the Single-Shot MultiBox Detector (SSD) in 2016, which was later refined to enhance performance [32]. For example, Souaidi et al. proposed the densely connected single-shot multi-box detector (DC-SSDNet), which improves small polyp detection [33]. Belabbes et al. developed the Small Polyp Detector (SPDNet) by reusing prediction layer information and integrating an attention mechanism [34].

The YOLO series has also gained attention for its efficiency in polyp detection. Pacal et al. applied YOLOv3 and YOLOv4 with data augmentation and Cross Stage Partial Network (CSPNet) [35]. Karaman et al. further optimized YOLOv4 using the artificial bee colony (ABC) algorithm for hyperparameter tuning [36]. Lalinia et al. employed data augmentation strategies with YOLOv8 [6]. Wang et al. integrated an efficient channel attention (ECA) mechanism and an interpretable Shapley explanation network (ShapNet) to enhance model interpretability in YOLOv5 [37]. Albuquerque et al. proposed a hybrid framework combining five detection models, achieving higher accuracy [38].

Despite the efficiency improvements of single-stage detection models, challenges remain in detecting small polyps and those with blurred boundaries, leading to missed diagnoses.

To overcome these challenges, this paper introduces HGNet. The next section provides a detailed description of the HGNet architecture.

## 3. Methods

### *3.1. The Research framework*

This paper proposes HGNet: High-Order Spatial Awareness Hypergraph and Multi-Scale Context Attention Network for colorectal polyp detection, systematically improved based on the YOLOv11 architecture. In the data preprocessing stage, image inversion, translation, and mosaic multi-image stitching strategies are applied. These enhancements improve the model's adaptability to scale variations and occlusion scenarios. The improved detection framework consists of a backbone network, neck network, and prediction head. Preprocessed images pass sequentially through these three modules for feature extraction, fusion, and detection tasks. As shown in Figure 1, the architecture of HGNet is illustrated.

The backbone network extracts low-level visual features of polyp images, such as color, shape, and texture. Its core structure includes standard convolution layers, C3K2 modules, Spatial Pyramid Pooling Fast (SPPF) modules, and the proposed Efficient Multi-scale Context Attention (EMCA) module. The C3K2 module performs feature transformation using input/output convolution layers and a Bottleneck structure with residual connections. The c3k parameter toggles the inter-layer connection mode between True and False. The SPPF module adjusts channel dimensions through $1\times1$ convolutions and concatenates three max-pooling layers to extract multi-scale features. The EMCA module, located at the end of the backbone network, uses a multi-branch attention mechanism. This mechanism fuses contextual semantic information from different receptive fields, enhancing the representation of polyp morphological changes.

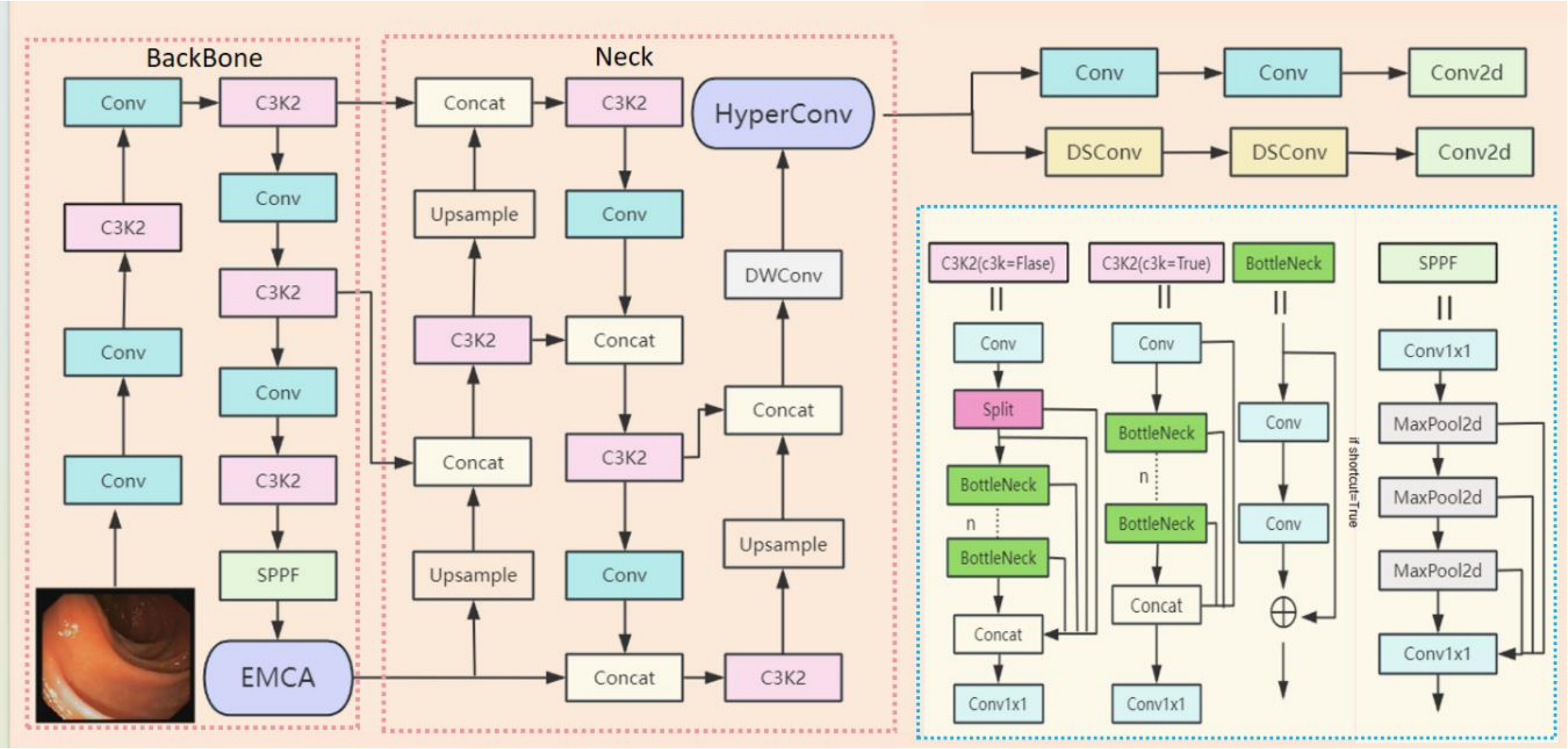

Figure 1 Architecture of the HGNet. The red box highlights the backbone network for feature extraction and the neck network for feature fusion. The top-right corner shows the detection head for localization and classification. The bottom-right blue box illustrates details of specific module.

The neck network builds upon the Feature Pyramid Network (FPN) architecture. First, it upsamples the output of the C3K2 module and cross-scales it with high semantic features from the previous layer. This achieves multi-level feature interaction. Then, Depthwise Convolution (DWConv) compress the channel dimensions to reduce computational redundancy.

A key innovation is the spatial hypergraph convolution module (HyperConv), which models higher-order feature relationships through hypergraph computations. Its functions include: (a) Generating hyperedges based on a preset feature distance threshold using Manhattan distance. Nodes with a feature distance smaller than the threshold are automatically grouped into cross-location hyperedges. This eliminates the need for predefined adjacency relationships and overcomes limitations of traditional CNN receptive fields and rigid adjacency definitions in Graph Convolutional Networks (GCNs); (b) Aggregating cross-scale contextual information through hyperedge aggregation. This mitigates feature loss caused by blurred polyp edges and reduces local feature degradation under mucosal folds and light reflection interference; (c) Enhancing the representation of small-scale polyp boundaries and cross-location semantic consistency. This improves the robustness of detecting complex morphological lesions.

The prediction head uses a dual-branch structure for polyp classification and localization. The classification branch consists of two standard convolution layers and a Conv2d layer, predicting the probability of lesion categories. The regression branch optimizes bounding box localization accuracy. It does so using two Depthwise-Separable Convolution (DSConv) and Conv2d layers, while reducing computational complexity. The structural design and implementation principles of the **Efficient Multi-scale Context Attention (EMCA) module** and the **Spatial Hypergraph Convolution (HyperConv) module** will be detailed in the following sections.

### *3.1.1. EMCA module*

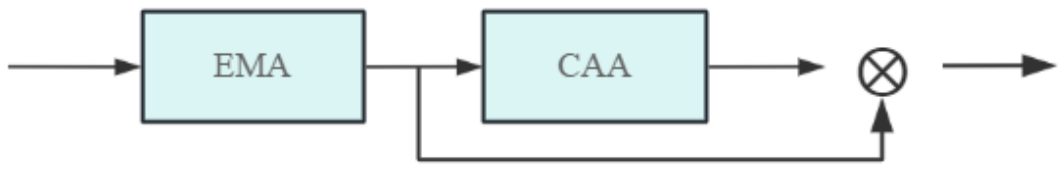

Figure 2 EMCA module

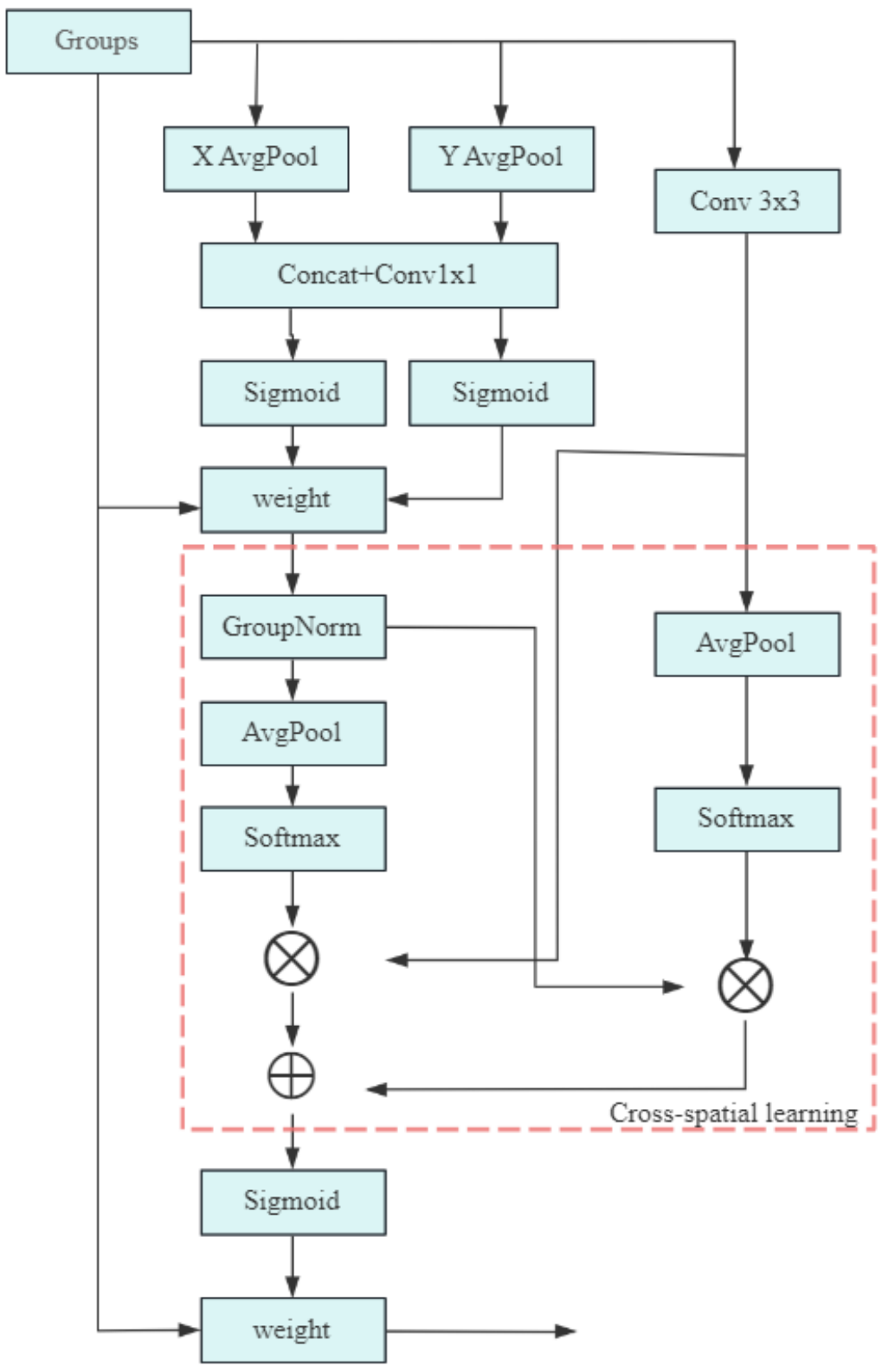

Figure 3 EMA module

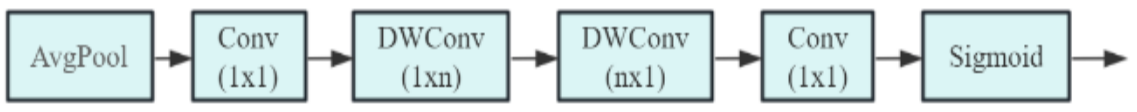

Figure 4 CAA module

To enhance feature extraction and emphasize critical polyp features, this paper introduces the Efficient Multi-Scale Context Attention (EMCA) module, as illustrated in Figure 2. EMCA sequentially integrates Efficient Multi-Scale Attention (EMA) and Context Anchor Attention (CAA).

a)The **EMA module** is illustrated in Figure 3. In this module, the input feature map $X_1 \epsilon R^{C\times H\times W}$ is initially divided into $g$ groups, resulting in a grouped feature map$X_g \epsilon R^{c/g\times H\times W}$ .This is then split into two branches: one branch performs global average pooling in both the horizontal and vertical directions to capture global information. Then the pooled features are using a 1×1 convolutional kernel to produce the feature map $X_2$, as shown in Eq. (1):

$$X_2 = Conv_{1\times 1}\left(Concat\left(pool_{h,}pool_{w,}\right)\right), \quad (1)$$

Subsequently, the feature map $X_2$ is passed through the Sigmoid activation function to generate the attention weights, which are then multiplied element-wise with the grouped feature map $X_g$ to obtain the weighted feature map. This is followed by group normalization to produce the feature map $x_1$, as shown in Eq (2):

$$x_1 = GN\left(\sigma(X_2)\otimes X_g\right), \quad (2)$$

where $GN$ denotes group normalization, $\sigma$ represents the Sigmoid activation function, and $\otimes$ denotes element-wise multiplication (Hadamard product).

The other branch is processed using a 3×3 convolutional kernel to generate the feature map $x_2$, as shown in Eq (3):

$$x_2 = Conv_{3\times 3}\left(X_g\right), \quad (3)$$

In the cross-spatial learning process, the feature maps $x_1$ and $x_2$ are separately subjected to global average pooling, followed by the application of the Softmax function to generate the feature maps $x_{11}$ and $x_{21}$. Then, element-wise multiplication is performed between $x_1$ and $x_{21}$ , as well as between $x_2$ and $x_{11}$ .The resulting spatial attention maps are subsequently summed, and the Sigmoid function is applied to normalize the values within the range [0, 1]. Finally, the attention coefficients are multiplied with the grouped feature map $X_g$ to produce the final weighted attention map $X_3$, as shown in Eq. (4):

$$x_{11} = \sigma_1\left(GAP(x_1)\right),$$
$$x_{11} = \sigma_1\left(GAP(x_1)\right), \quad (4)$$
$$X_3 = \sigma_2(x_{11}\otimes x_2 + x_{21}\otimes x_1)\otimes X_g,$$

where $\sigma_1$ is the Softmax function, $\sigma_2$ is the Sigmoid function, and $\otimes$ denotes element-wise multiplication.

b)The **CAA module**, as shown in Figure 4, operates as follows: the input feature map $X_3 \in R^{C\times H\times W}$ undergoes a 7×7 average pooling operation to capture global information. Then, the number of channels is adjusted using a 1×1 convolution to obtain the feature map $F_1$, as expressed in Eq. (5).

$$F_1 = Conv_{1\times 1}(AvgPool_{7\times 7}(X_3)), \quad (5)$$

Subsequently, a DWConv is applied separately in the horizontal and vertical directions to capture directional information. Afterward, a 1×1 convolution is used to adjust the number of channels, followed by the application of the Sigmoid activation function to

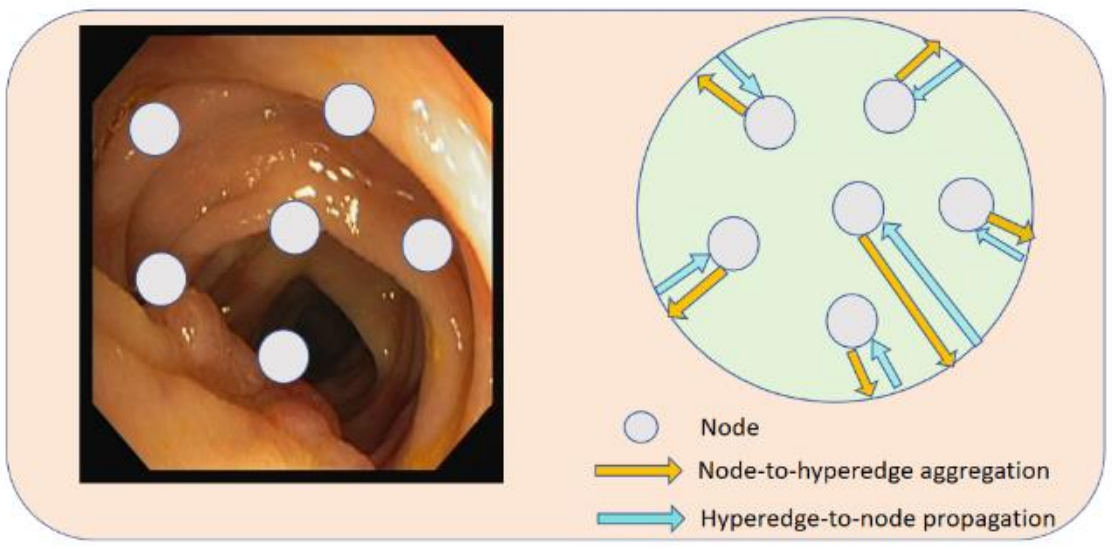

Figure 5 Learning process on the HyperConv

obtain the attention coefficients. This process is expressed in Eq. (6):

$$F_2 = DWConv_{k\times 1}\left(DWConv_{1\times k}(F_1)\right), \quad F_3 = \sigma\left(Conv_{1\times 1}(F_2)\right), \tag{6}$$

where $\sigma$ is a Sigmoid function.

Finally, the size of the attention coefficients is adjusted, and they are element-wise multiplied with the original input feature map to obtain the weighted feature map.

### *3.1.2. HyperConv module*

This paper introduces a Spatial Domain Hypergraph Convolution (HyperConv) module, which is deployed at the frontend of the detection head to facilitate feature aggregation. The hypergraph structure is mathematically represented as $\mathcal{G} = \{\mathcal{V}, \mathcal{E}\}$, where $\mathcal{V}$ denotes the set of nodes and $\mathcal{E}$ represents the set of hyperedges. The construction process follows a systematic procedure:

Initially, the input feature map $X\epsilon \mathrm{R}^{C\times H\times W}$ undergoes a gridization process. Each pixel within the feature map is treated as a node $v_i\epsilon\mathcal{V}$, where the feature dimension of each node is $C$, and the total number of nodes is $N = H \times W$. The next step involves calculating the distance between any two nodes $v_i$ and $v_j$ using the Manhattan distance metric. Specifically, if the distance $\parallel x_j - x_i \parallel_1$ between two nodes is smaller than a predefined threshold $\delta$, a hyperedge is formed between nodes $v_i$ and $v_j$. Thus, for any given node $v_i\epsilon\mathcal{V}$, the corresponding hyperedge set $e_i$ is expressed in Eq. (7):

$$e_i = \{v_j \mid v_j \,\epsilon \mathcal{V},\ \parallel x_j - x_i \parallel_1 < \delta\}, \tag{7}$$

Moreover, the associated hyperedge set for each node $v_j$ is as shown in Eq. (8):

$$\mathcal{N}_e(v_j) = \{e_i \mid v_j \epsilon e_i, v_j \epsilon \mathcal{V}\}, \tag{8}$$

As depicted in Figure 5, the HyperConv operates in two stages to propagate the features. The first stage involves message aggregation from nodes to hyperedges. In this stage, the hyperedge information $X'$ is computed via the adjacency matrix $H\epsilon R^{|\mathcal{V}|\times|\mathcal{E}|}$, as shown in Eq. (9) [39]:

$$X' = D_e^{-1} H^T X \theta, \tag{9}$$

where $\theta$ represents the learnable parameters, and $D_e \epsilon R^{|\mathcal{E}|\times|\mathcal{E}|}$ is the hyperedge degree matrix. The diagonal elements of $D_e$ correspond to the number of nodes connected to each hyperedge. The second stage is the message propagation from hyperedges back to the nodes, which results in the updated node features $X''$, expressed in Eq. (10) [39]:

$$X'' = D_v^{-1} H X', \tag{10}$$

where $D_v \epsilon R^{|\mathcal{V}|\times|\mathcal{V}|}$ denotes the node degree matrix, with its diagonal elements representing the number of hyperedges connected to each node.

## *3.2. The Loss function*

The loss functions include classification loss ($L_{cls}$), bounding box regression loss ($L_{box}$), and distributed focus loss ($L_{dfl}$). It is expressed in Eq (11):

$$Loss = \lambda_1 L_{cls} + \lambda_2 L_{box} + \lambda_3 L_{dfl}, \tag{11}$$

where $\lambda_1$, $\lambda_2$, $\lambda_3$ represent the weights for each type of loss, respectively.

Classification loss is expressed using binary cross-entropy loss, as shown in Eq (12):

$$L_{cls} = -\sum_{i=1}^{N} [y_i . log(p_i) + (1 - y_i) . log(p_i)], \tag{12}$$

where $p_i$ is the predicted probability of the $i - th$ class, and $y_i$ is the true class label, which can be either 0 or 1.

The bounding box regression loss ($L_{box}$) is shown in Eq (13):

$$L_{box} = 1 - CLOU,$$
$$CLOU = IoU - \frac{d^2}{c^2} - \frac{\alpha^2}{1 - IoU + \alpha},$$
$$IoU = \frac{A \cap A_{gt}}{A \cup A_{gt}}, \tag{13}$$
$$\alpha = \frac{4}{\pi^2}\left(tan^{-1}\left(\frac{w_{gt}}{h_{gt}}\right) - tan^{-1}\left(\frac{w}{h}\right)\right)^2,$$

where $IoU$ denotes the intersection over union between the predicted and ground truth bounding boxes, $d$ and $c$ represent the distance from the predicted box to the ground truth box and the diagonal length of the smallest enclosing bounding box, respectively, and $\alpha$ denotes the difference in aspect ratio.

Distributed focus loss ($L_{dfl}$) is determined by the probability distribution of the predicted bounding box position $[y_n, y_{n+1}]$, which directs the model to focus more on bounding boxes closer to the ground truth. As shown in Eq (14):

$$L_{dfl} = (y - y_n)\, log(S_{n+1}) - (y_{n+1} - y)\, log(S_n),$$
$$S_n = \frac{y_{n+1} - y}{y_{n+1} - y_n}, \quad (14)$$
$$S_{n+1} = \frac{y - y_n}{y_{n+1} - y_n},$$

where $S_n$ and $S_{n+1}$ denote the probabilities of the predicted bounding box being located at the two discrete positions $y_n$ and $y_{n+1}$, respectively.

## 4. Experiments

### 4.1. Datasets

The dataset used is from the 8th National Biomedical Engineering Innovation Design Competition of China (2023), comprising 28,773 colorectal polyp images at various resolutions, categorized as hyperplasia and adenoma. Due to computational limitations, 4,000 images were randomly selected from the original dataset to construct a subset, named BpolypD. Additionally, the Kvasir-SEG [40] and CVC-ClinicDB [41] datasets, sourced from Kaggle, were utilized for model evaluation. The former includes 1,000 images with resolutions ranging from 332×487 to 1920×1072, while the latter contains 612 polyp images at a resolution of 384×288. All datasets were split into training, validation, and test sets at an 8:1:1 ratio.

### 4.2. Experiment settings

This paper is based on the PyTorch framework for all experiments, conducted on a Windows 11 system with an Intel(R) Xeon(R) W-2255 CPU, 64GB of memory, and an NVIDIA GeForce RTX 3070 GPU. The PyTorch version is 2.0.0, with CUDA 11.7 support and Python 3.8.20. Other model-related parameter settings are shown in Table 1.

Table 1
Model Parameter Settings

| Parameter | Description | Value |
|---|---|---|
| $lr0$ | initial learning rate | 0.01 |
| $lrf$ | final learning rate | 0.01 |
| $epoach$ | training iteration | 40 |
| $\lambda_1$ | the weight of classification loss | 3 |
| $\lambda_2$ | the weight of bounding box loss | 4 |
| $\lambda_3$ | the weight of distributed focus loss | 1.5 |

### 4.3. Evaluation Metrics

To evaluate the model's performance in polyp detection, this study uses standard object detection metrics: precision, recall, F1-Score, and mean Average Precision. These metrics are based on True Positives ($TP$), False Positives ($FP$), and False Negatives ($FN$). A predicted bounding box is considered a $TP$ if its $IoU$ with the ground truth exceeds a set threshold; otherwise, it is considered an $FP$. If no predicted bounding box exceeds the $IoU$ threshold for any remaining ground truth, it is defined as an $FN$.

Precision ($P$) is the proportion of correctly identified positive samples among all predicted positives, as shown in Eq (15):

$$P = \frac{TP}{TP + FP}, \quad (15)$$

Recall ($R$) is defined as the ratio of correctly identified positive samples to the total number of actual positive samples, as expressed in Eq. (16):

$$R = \frac{TP}{TP + FN}, \quad (16)$$

The F1-score ($F1$) is the harmonic mean of precision and recall, with values in the [0,1] range. A higher value indicates better detection performance, as shown in Eq (17):

$$F1 = \frac{2 \times P \times R}{P + R}, \quad (17)$$

The mean average precision ($mAP$) measures the model's performance across all categories at different confidence thresholds, with values ranging from 0 to 1., as shown in Eq (18):

Table 2
Comparative results of different models

| | 模型 | P | R | F1 | mAP@0.5 |
|---|---|---|---|---|---|
| BpolypD | YOLOv9 | 0.875 | 0.919 | 0.896 | 0.950 |
| | YOLOv10n | 0.871 | 0.840 | 0.855 | 0.919 |
| | YOLOv11 | 0.920 | 0.892 | 0.907 | **0.963** |
| | SSD | 0.814 | 0.708 | 0.751 | 0.818 |
| | Faster R-CNN | 0.673 | **0.925** | 0.775 | 0.926 |
| | HGNet | **0.940** | 0.906 | **0.923** | 0.960 |
| Kvasir-SEG | YOLOv9 | **0.874** | 0.800 | 0.835 | 0.855 |
| | YOLOv10n | 0.800 | 0.760 | 0.779 | 0.790 |
| | SSD | 0.872 | 0.394 | 0.540 | 0.782 |
| | Faster R-CNN | 0.546 | 0.793 | 0.650 | 0.746 |
| | HGNet | 0.811 | **0.909** | **0.857** | **0.915** |
| CVC-ClinicDB | YOLOv9 | 0.776 | 0.507 | 0.613 | 0.595 |
| | YOLOv10n | 0.624 | 0.658 | 0.641 | 0.583 |
| | SSD | **0.855** | 0.296 | 0.440 | 0.662 |
| | Faster R-CNN | 0.603 | 0.508 | 0.550 | 0.505 |
| | HGNet | 0.804 | **0.767** | **0.785** | **0.812** |

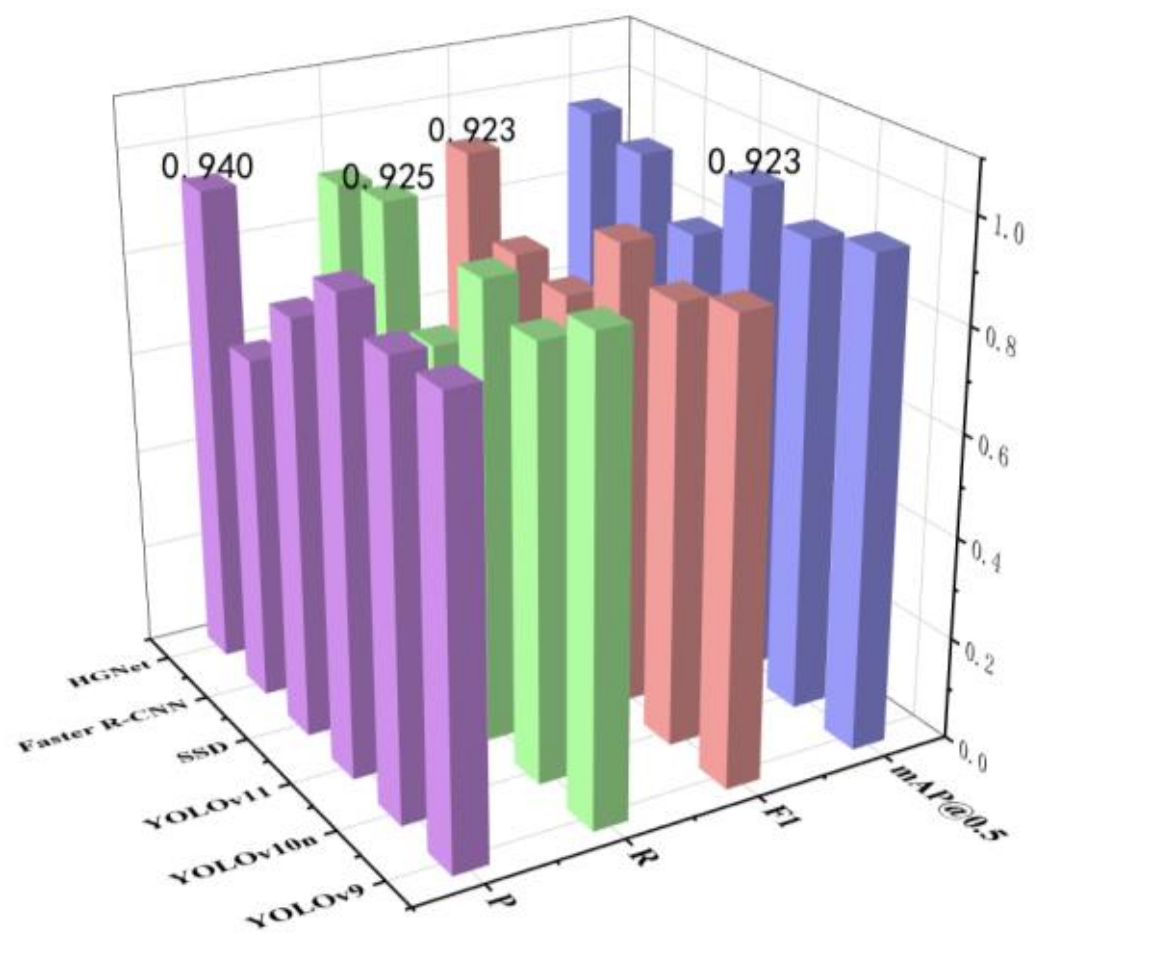

Figure 6 Model comparison results on the BpolypD dataset

$$mAP = \frac{\sum_{i=1}^{k} AP_i}{n}, \qquad (18)$$

where n denotes the total number of categories, and $AP_i$ represents the average precision ($AP$) value for the $i-th$ category.

$AP$ represents the area under the the precision-recall ( $P-R$ ) curve, reflecting the model's accuracy across individual categories at various confidence levels. as shown in Eq (19):

$$AP = \int_0^1 p(r)\, dr, \qquad (19)$$

## 5. Results

### *5.1. Comparsion*

To evaluate model performance, five object detection architectures were compared under identical experimental conditions: SSD [32], Faster R-CNN [28], YOLOv9 [42], YOLOv10n [43], and YOLOv11 [44]. To address the generalization challenges posed by the small-scale Kvasir-SEG and CVC-ClinicDB datasets, a transfer learning approach was employed. Specifically, models were initialized with weights pre-trained on the BpolypD dataset before being fine-tuned on the target datasets. Quantitative comparisons are provided in Table 2, with bold and underlined

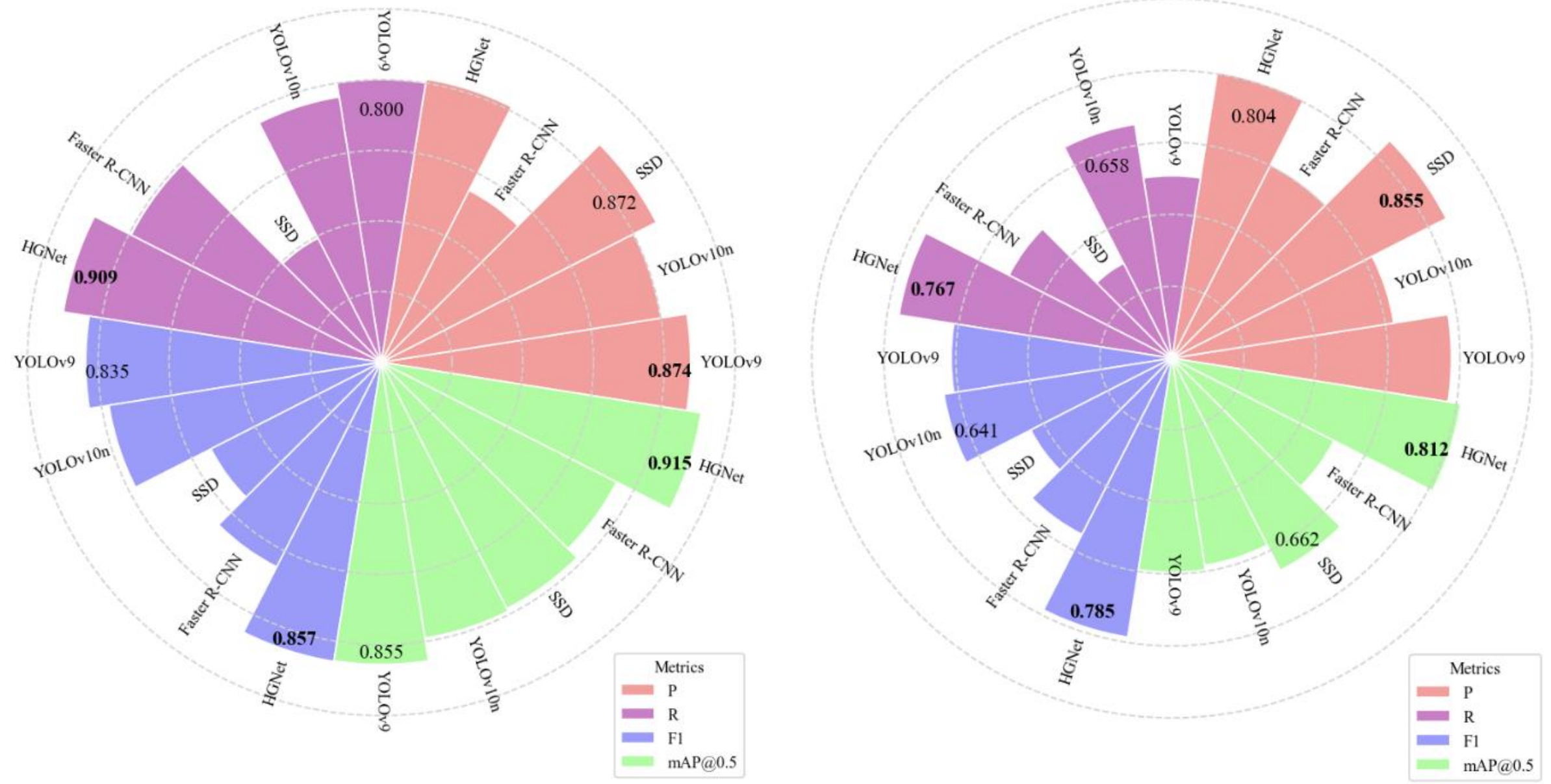

a b

Figure 7 presents model comparison results on the Kvasir-SEG (a) and CVC-ClinicDB (b) datasets values indicating the best and second-best performers, respectively.

Next, this study systematically presents the experimental results for each dataset using a three-dimensional visualization approach.

As shown in Figure 6, experiments on the BpolypD dataset demonstrate that HGNet achieves optimal performance, with an accuracy of 94.0% and an F1 score of 92.3%, surpassing the second-best model YOLOv11 by 2.0% and 1.6%, respectively. HGNet's recall of 0.906 ranks third, just 1.9% below the top model, Faster R-CNN. For mAP@0.5, HGNet scores 0.960, 0.3% lower than YOLOv11. Compared to the original YOLOv11 baseline, HGNet improves accuracy, recall, and F1 score by 2.0%, 1.4%, and 1.6%, respectively. These results highlight the effectiveness of the architectural modifications in enhancing performance to reach state-of-the-art levels.

As shown in Figures 7a and 7b, in the model comparison on the Kvasir-SEG and CVC-ClinicDB datasets, HGNet achieves optimal performance in recall, F1 score, and mAP@0.5 metrics. A detailed analysis is as follows:

**Kvasir-SEG dataset**: HGNet improves recall, F1 score, and mAP@0.5 by 10.9%, 2.2%, and 0.6%, respectively, compared to the second-best model, YOLOv9n. However, its accuracy is 6.3% lower than that of the best model, YOLOv9.

**CVC-ClinicDB dataset**: HGNet's recall and F1 score are 10.9% and 14.4% higher, respectively, compared to the second-best model, YOLOv10n. The mAP@0.5 is 15% higher than that of the second-best model, SSD, while its accuracy lags 5.1% behind the best model, SSD.

In summary, the HGNet model outperforms all comparison models in accuracy and F1 score on the BpolypD dataset, with improvements in accuracy, recall, and F1 score over the YOLOv11 baseline. Cross-dataset transfer learning tests show a slight accuracy decrease on Kvasir-SEG and CVC-ClinicDB, but HGNet excels in other metrics. This performance is attributed to the HyperConv module, which aggregates similar features and uses multi-scale context modeling to reduce noise, enhancing robustness in complex shape detection. The accuracy gap is linked to interference from low-correlation features. Overall, HGNet delivers superior performance with potential for further optimization.

a

b

c

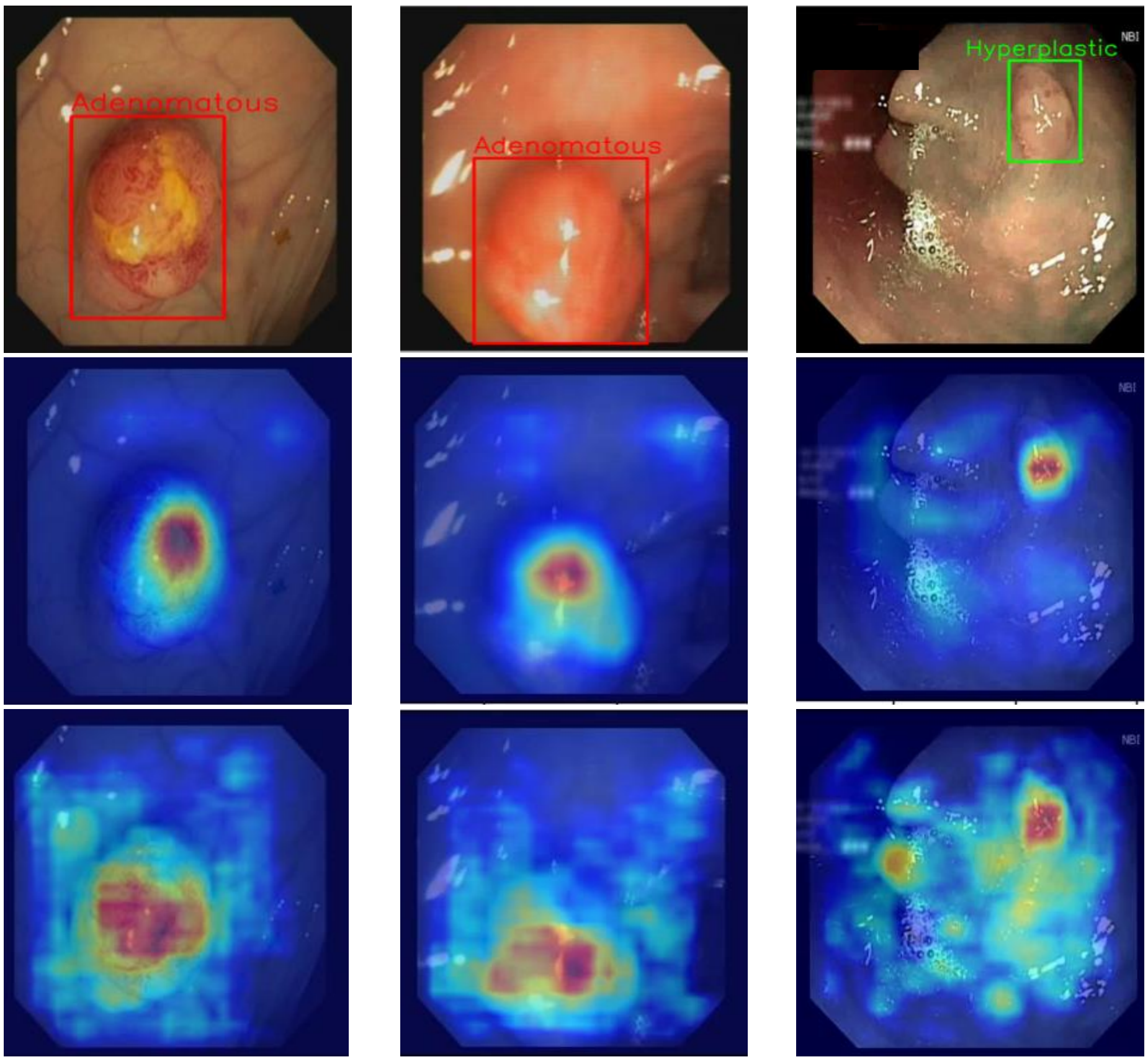

Figure 8 Eigen Eigen-CAM heatmap visualization on the BpolypD dataset. (a) Original images with annotations; (b, c) Heatmap visualizations from YOLOv11 and HGNet, respectively. Darker red regions indicate higher model attention, with (c) demonstrating better alignment with polyp morphology.

*5.2. Visualization result analysis*

This study utilizes the Eigen Class Activation Map (Eigen-CAM) to visualize the decision-making process of HGNet. The red intensity gradient represents the model's attention distribution. As shown in Figure 8, subfigure (a) presents two hyperplastic polyps and one adenomatous polyp with annotations.

Experimental results reveal that YOLOv11 (subfigure b) fails to capture some polyp regions, while HGNet (subfigure c) aligns more closely with the annotated morphology. HGNet occasionally activates non-polyp regions. This may result from its hypergraph convolution module, which aggregates long-range dependencies and enhances structural representation. Heatmap analysis shows that HGNet's activation boundaries better match clinical annotations, demonstrating higher accuracy and robustness in polyp recognition.

*5.3. Ablation experiments*

*5.3.1. Performance Analysis of Similarity Metrics in Hypergraph Construction*

Table 3 presents the impact of different similarity metrics on the performance of the hypergraph convolution module. The best and second-best results are highlighted in bold and underlined, respectively. To ensure fairness, all methods were evaluated under the same network architecture.

Table 3
Hypergraph Construction Comparison with Distance Metrics

| Distance Metric | P | R | F1 | mAP@0.5 | mAP@0.5-0.95 |
|---|---|---|---|---|---|
| Manhattan Distance | 0.938 | **0.901** | **0.919** | **0.941** | **0.670** |
| Euclidean Distance | **0.941** | 0.859 | 0.899 | 0.938 | 0.646 |
| Chebyshev Distance | 0.814 | 0.759 | 0.785 | 0.833 | 0.525 |
| Gaussian Kernel Function | 0.835 | 0.842 | 0.838 | 0.881 | 0.590 |

Table 4
Ablation Study of Modules

| Group | EMCA | HyperConv | P | R | F1 | mAP@0.5 | mAP@0.5-0.95 |
|---|---|---|---|---|---|---|---|
| 1 | | | **0.941** | 0.887 | 0.913 | 0.942 | 0.667 |
| 2 | √ | | 0.916 | **0.927** | 0.921 | 0.957 | 0.688 |
| 3 | | √ | 0.938 | 0.901 | 0.919 | 0.941 | 0.670 |
| 4 | √ | √ | 0.940 | 0.906 | **0.923** | **0.960** | **0.699** |

Experimental results show that Manhattan distance achieves the highest values across four key metrics: recall, F1-score, mAP@0.5, and mAP@0.5–0.95. Compared to the second-best Euclidean distance, it improves these metrics by 4.2%, 2.0%, 0.3%, and 2.4%, respectively, while maintaining a negligible 0.3% difference in accuracy.

From a computational perspective, Euclidean distance involves squared operations, and Gaussian similarity relies on exponential functions, whereas Manhattan distance only requires absolute value computations. This enables the construction of a sparse hypergraph adjacency matrix with significantly improved efficiency. Moreover, Manhattan distance demonstrates greater robustness in noisy images. These findings validate its superiority in both polyp localization and classification accuracy.

#### *5.3.2. Effect of of Different Modules*

This study quantitatively evaluates the impact of different modules on detection performance through ablation experiments. The results are shown in Table 4. Introducing the Efficient Multi-scale Context Attention (EMCA) module alone increases recall by 4.0%, with improvements in $F1\ score$, $mAP@0.5$, and $mAP@0.5-0.95$, but accuracy decreases by 2.5%. The Hypergraph Convolution (HyperConv) module alone boosts recall by 1.4%, with gains in $F1\ score$ and $mAP@0.5-0.95$, while accuracy and $mAP@0.5$ drop slightly by 0.3% and 0.1%. When both modules are combined, recall increases by 1.9%, $F1\ score$ by 1.0%, and $mAP@0.5$ by 1.8%, with accuracy loss reduced to 0.1%.

These results show that the EMCA module enhances recall through multi-scale context fusion, and the HyperConv module improves robustness by modeling higher-order spatial relationships. The combined modules optimize multiple metrics with minimal accuracy loss.

### *5.4. Dissussion*

The HGNet model extracts multi-scale features using the Efficient Multi-scale Contextual Attention (EMCA) module. Its parallel branch structure improves feature representation by integrating contextual information across different scales. The spatial-domain-based HyperConv module establishes spatial relationships between cross-position nodes through hyperedges, optimizing feature fusion via high-order semantic relationships.

Experiments on the BpolypD dataset show that HGNet outperforms the YOLOv11 baseline in accuracy and F1-score. Transfer learning helps address the issue of limited training samples, leading to improved performance in most metrics. However, accuracy decreases in some cases, indicating that the current hypergraph construction method is limited by the generalization capability of local features. Further refinement of the hypergraph construction is needed.

Eigen-CAM heatmap analysis shows that the model's attention distribution closely matches the true anatomical structure of polyps. Minor misattention in non-polyp regions is observed, but these signals help distinguish blurred boundary areas. Ablation studies confirm the effectiveness of the modules. The EMCA module enhances semantic expression by integrating cross-scale features, while the HyperConv module improves feature consistency through high-order spatial relationships. The interaction between these modules significantly boosts performance.

## 6. Conclusion

This paper introduces the High-Order Spatial Awareness Hypergraph and Multi-Scale Context Attention Network (HGNet) for detecting small colorectal polyps. HGNet integrates an Efficient Multi-scale Context Attention (EMCA) module to extract key polyp features and a hypergraph convolution (HyperConv) module to capture high-order node relationships. Transfer learning strategies address the challenge of limited training data, and heatmap visualizations help interpret the model's predictions. Experimental results demonstrate that HGNet outperforms most comparison models, showcasing superior detection capability and robustness. Future work will focus on refining hypergraph construction and network design to enhance detection accuracy and recall, providing better support for clinical polyp diagnosis.

## CRediT authorship contribution statement

**Xiaofang Liu**: Writing – original draft, Conceptualization, Data curation, Formal analysis, Methodology, Software. **Lingling Sun**: Funding acquisition, Resources, Validation. **Xuqing Zhang**: Funding acquisition, Supervision, Visualization. **Bin Zhao**: Writing – review & editing, Funding acquisition, Methodology, Resources, Supervision.

## Declaration of competing interest

The authors declare that they have no known competing financial interests or personal relationships that could have appeared to influence the work reported in this paper.

## Acknowledgments

This work was supported by the Guizhou Provincial Basic Research Program (Natural Science Category), Project ZK[2023]298, MS[2025]557, China. The authors would also like to express their sincere gratitude to the editor and reviewers for their valuable feedback and constructive comments.

## Data availability

The datasets used in this study, including BPolypD, Kvasir-SEG, and CVC-ClinicDB, are publicly available:

- **BPolypD**: https://aistudio.baidu.com/datasetdetail/216022.
- **Kvasir-SEG**: https://datasets.simula.no/kvasir-seg/.
- **CVC-ClinicDB**: https://www.kaggle.com/datasets/balraj98/cvcclinicdb.